# Aspect-Oriented Summarization for Psychiatric Short-Term Readmission Prediction

**WonJin Yoon** [1,4] **Boyu Ren** [2,4] **Spencer Thomas** [1] **Chanhwi Kim** [3]
**Guergana Savova**[1,4] **Mei-Hua Hall** [2,4] **Timothy Miller** [1,4]
[1] Boston Children's Hospital, MA, USA [2] McLean Hospital, MA, USA
[3] Korea University, South Korea [4] Harvard Medical School, MA, USA
{wonjin.yoon, spencer.thomas, guergana.savova, timothy.miller}@childrens.harvard.edu
bren@mgb.org chanhwi_kim@korea.ac.kr mhall@mclean.harvard.edu

## Abstract

Recent progress in large language models (LLMs) has enabled the automated processing of lengthy documents even without supervised training on a task-specific dataset. Yet, their zero-shot performance in complex tasks as opposed to straightforward information extraction tasks remains suboptimal. One feasible approach for tasks with lengthy, complex input is to first summarize the document and then apply supervised fine-tuning to the summary. However, the summarization process inevitably results in some loss of information. In this study we present a method for processing the summaries of long documents aimed to capture different important aspects of the original document. We hypothesize that LLM summaries generated with different aspect-oriented prompts contain different *information signals*, and we propose methods to measure these differences. We introduce approaches to effectively integrate signals from these different summaries for supervised training of transformer models. We validate our hypotheses on a high-impact task – 30-day readmission prediction from a psychiatric discharge – using real-world data from four hospitals, and show that our proposed method increases the prediction performance for the complex task of predicting patient outcome.

## 1 Introduction

Recent progress in large language models (LLMs) has allowed the processing of long documents that were previously difficult to process due to limitations in the model sequence length (Zeng et al., 2024). Although this enabled a variety of NLP tasks to be applied to longer documents, particularly using a zero- or few-shot approach, some tasks that require deeper text processing beyond information extraction remain challenging in zero-shot settings (Yoon et al., 2024; Fan et al., 2024). One such complex task is patient outcome predictions such as out-of-hospital mortality or readmission prediction, where the task objective is to predict a future event given the summary of the patient's stay as recorded at their discharge in a type of clinical note called a *discharge note*. In this work, we are concerned with the high-impact real-world task of predicting readmissions in psychiatric hospitals.

Due to the immense computational cost required to fine-tune LLMs (especially given the length of clinical documents), and the regulatory challenges presented by transferring sensitive patient data to a large compute environment, traditional supervised training approaches are not feasible for tasks where patient data are the primary source.[1] On the other hand, recent studies show that one of the core LLM strengths is their ability to generate high-quality summaries (Zhang et al., 2024b; Liu et al., 2023; Van Veen et al., 2024). Furthermore, clinical notes contain detailed information about the patient's disease/s not all necessarily relevant to a particular classification task (Hultman et al., 2019). Thus, summarization emerges as a feasible approach to retain only the relevant content.

The recent advances in LLM capabilities make it possible to explore an approach where long documents are first summarized to an acceptable length, and the summaries are used to fine-tune a smaller language model. This approach takes advantage of the strengths of LLM while avoiding the challenges of fine-tuning them (Chen et al., 2024). However, since the summarization process shortens a document, some content details that are important signals for downstream tasks might be removed. Aspect-oriented prompting (Ahuja et al., 2022), where prompt variations are used to condition the summary on important aspects of the text (e.g., risk factors), could capture the relevant details more reliably.

[1]Due to the Transformers architecture, a vanilla method to fine-tune a model with $n$ context size will require $O(n^2)$ of GPU memory.

Therefore, we address the following research questions: 1) Does the use of LLM-based aspect-oriented summarization extract measurably different information signals with different information-focused prompts (i.e. aspects)? 2) What are effective strategies for merging signals from different aspect-oriented summaries?

The present work makes the following contributions to answering these research questions. First, we develop and quantify methods for measuring the information signal differences in the summarized text. Working with our domain experts, we created three different types of prompts to summarize discharge notes from the Electronic Health Record (EHR) into a paragraph-length document. We generated summarized documents using the prompts and used them as training data for the downstream task of readmission prediction by fine-tuning smaller pre-trained language models. During this process, the differences in information contained in the summaries become internalized within the fine-tuned models, allowing to measure these differences by comparing the prediction differences. For scientific rigor, we fine-tuned repeatedly with random seed variations, setting the control group as the variation among models trained using the same prompt and the experimental group as the variation among models trained using different prompts.

Second, we explore methods for combining summaries generated by different aspect-oriented prompts and propose their integration at the dataset level. We show that our proposed method improves the performance of fine-tuned models and demonstrate that effectively integrating information from diverse summaries generated via aspect-oriented summarization yields better performance than fine-tuning with a single summary.

## 2 Dataset and task

In this section, we describe the readmission prediction task and the data sourced from real-world hospital EHRs.

### 2.1 Source of experimental data

We extracted psychiatric discharge notes from the EHR databases of four hospitals within a single academic health center in the United States. Patient encounters were selected with two criteria: (1) range for patient's age at admission of 18-65 years old; pediatric and elderly patients excluded because of the disease specifics for these age groups, and (2) ICD-10 diagnosis codes for mood disorders or psychotic disorders (All codes starting with F2 or F3). Table 1 shows the number of notes in the training, development, and test sets grouped by hospital. This study was approved by the Mass General Brigham Institutional Review Board with a formal reliance at Boston Children's Hospital. All experiments, including data processing, were conducted within the secure firewall of the health center in HIPAA-compliant environments.

| Hospital | Train | Dev | Test | Total | Pos/Total |
|---|---|---|---|---|---|
| McLean | 6066 | 775 | 1718 | 8559 | 0.2914 |
| MGH | 8840 | 1392 | 2610 | 12842 | 0.2979 |
| BWH | 979 | 130 | 339 | 1448 | 0.3384 |
| FH | 793 | 94 | 242 | 1129 | 0.2214 |

Table 1: Dataset distribution by hospital sites. The middle section lists the note counts for the train, development (dev), and test splits; the right section shows the positive label ratio (Pos/Total).

### 2.2 Task definition

The psychiatric short-term readmission prediction task involves determining whether a patient will be readmitted to the hospital within **30 days after discharge**. This task is important for the patient's quality of life (Owusu et al., 2022; Ren et al., 2025) as well as used by the United States Centers for Medicare and Medicaid Services (CMS) as a quality metric for some conditions that is tied to reimbursement rates (CMS).

This short-term readmission prediction task differs from long-term or lifetime readmission prediction in nature; a more detailed discussion of this distinction is provided in Appendix J.

The dataset's unit is a hospital admission, and a single patient might have multiple admissions. Since discharge notes are generated once per admission at the time of discharge, there is a one-to-one correspondence between a discharge note and a hospital admission, making both units equivalent in the dataset. To ensure that discharge notes from the same patient do not scatter across the training, development and test splits, we used the patient ID. In this study, the model's input is the text of a psychiatric discharge note, without any EHR structured data, i.e. ICD-10 codes or medication orders. The label is binary: a positive label indicates readmission within 30 days, while a negative label indicates no readmission within that period.

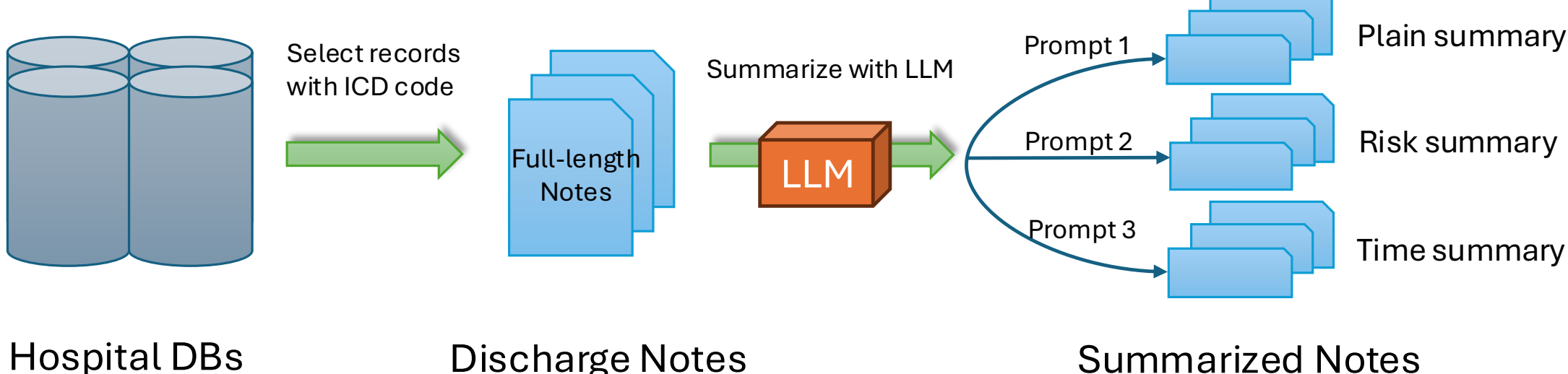

Figure 1: Overview of our aspect-oriented summarization pipeline. Discharge notes sourced from hospitals' EHR databases and processed with LLM prompts. The summarized notes are used in the experiments detailed in Sections 3.2 and 3.3.

## 3 Methods

Our methodology comprises of three components: aspect-oriented summarization, measuring information signal differences, and integrating information signals with a focus on readmission prediction.

### 3.1 Aspect-oriented summarization

The intrinsic nature of summarization leads to information loss as some of the original content is omitted from the summary (Tang et al., 2023). Summarization becomes challenging if the document is long and if the source document is written in a nuanced language, where overlooking even small details could significantly alter the intended meaning (Zhang et al., 2024a).

Psychiatric discharge notes are valuable resources for assessing LLM summarization capabilities, not only due to their length but also because of their nuanced language. A plain-language description of an event could be a significant signal. For example, an event where a patient stops taking a medication is informative, but its importance and meaning is highly dependent on context – whether it was due to side effects, ineffectiveness, or some other reason.

These factors motivate our aspect-oriented summarization approach. Data processing steps are shown in (Figure 1). We started with the raw EHR data and extracted the discharge notes. These discharge notes are summarized using an instruction-tuned LLM. Three types of prompts, `plain`, `riskfactor`, and `timeline`, were used to produce different types of summaries, i.e. aspects. The prompt templates are shown in Figure 2 and in Appendix A. The `plain` prompt was intended to create a generic summary; the `riskfactor` prompt focused on specific research factors previously published in psychiatric NLP research (Holderness et al., 2019; Ding et al., 2020); and the `timeline` prompt was designed to generate summaries containing an ordered sequence of important events before and during the admission. All three prompts were developed with input from researchers with clinical NLP and psychiatric expertise.

We use the summarized discharge notes in the experiments presented in Sections 3.2 and 3.3.

### 3.2 Measuring information differences

Since we hypothesize that different types of summaries potentially contain non-overlapping pieces of information, we describe our method to quantify the degree of overlap.

Existing methods for automatically comparing documents, such as ROUGE (Lin, 2004) and BERTscore (Zhang et al.), are designed for purposes not fitting our goal to estimate the information differences of summaries intended to represent different sides of the original document. Moreover, they are inadequate for capturing subtle information differences. Based on our preliminary study, manually evaluating summaries for information preservation, i.e., determining whether important information is kept or omitted, is not scalable and is subjective. Therefore, we developed a data-driven approach.

The intuition behind our approach is that models capture signals specific to the text used during the supervised training phase. Therefore, if during the supervised training stage the model is exposed to aspect-oriented summaries each targeting a different side of the original text, the model will learn the various aspects. Therefore, we train task-specific models on each type of summary, expecting the models to learn the summary-specific signals which in turn will be reflected in the final predictions. By comparing the outputs of the task-specific models we measure the difference in the signals present in the inputs. Figure 3 is the Venn diagram illustrating that models derived from different summaries

**Plain summarization prompt**

<s>[INST] The following is a discharge summary for a patient leaving a psychiatric hospital.
***Discharge note inserted here.***
This is the end of the discharge summary. Please summarize the patient's hospital course in a paragraph. [/INST]

**Timeline-focused summarization prompt**

<s>[INST] The following is a discharge summary for a patient leaving a psychiatric hospital.
***Discharge note inserted here.***
This is the end of the discharge summary. Write a very brief semi-structured timeline with the major events in the patient's medical history, including prior to admission. Do not give a full medication history, though recent medications and a note of a complex medication history are useful. Each line has the form <Date/Time>: <Event description> [/INST]

Figure 2: Prompts for summarization. Three types of prompts, `plain`, `riskfactor`, and `timeline`, were used to generate different types of summaries. For `riskfactor` prompt, please see Appendix A

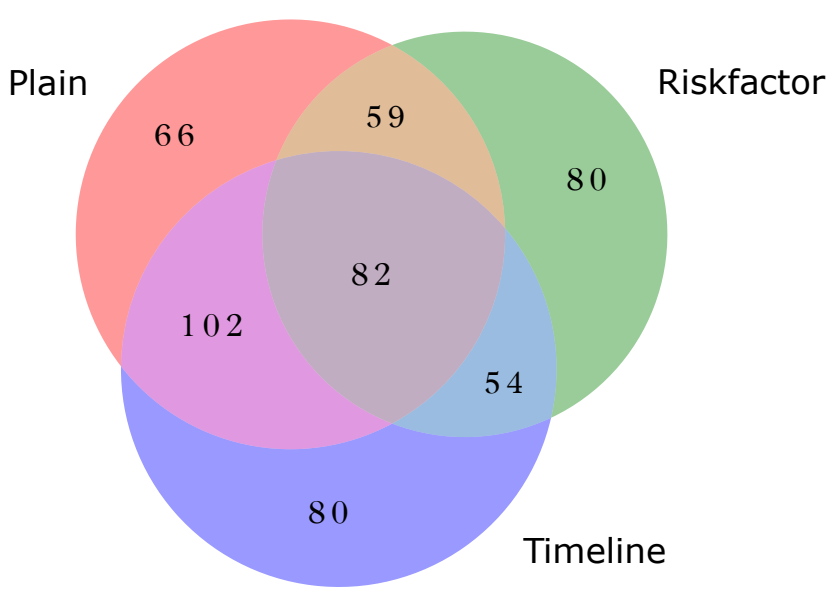

Figure 3: Differences between predictions (True Positives) of task-specific models fine-tuned with different summary types. The non-overlapping predictions suggest that aspect-oriented prompting extracts different yet complimentary information from the source document. The figure shows experiments from MGH with a random seed set to 0.

make distinct decisions.[2]

We iterated model training with identical settings but with different seeds on the same summary types. Intra-aspect experiments (i.e. measuring the difference between summaries generated using the same prompt) were conducted as a control group and compared against inter-aspect experiments (i.e. measuring the difference between summaries generated using different prompts). This approach aims to mitigate differences arising from sources of randomness such as random initialization.

#### 3.2.1 Pair-wise similarity scores between models

We assess the similarity between the outputs of any given pair of models. The outputs were ranked based on probabilities, and Kendall's tau was employed to evaluate the similarity between the two ranked lists.[3]

We define a list of prediction from a model $M_1$ on $k$ notes as $L_1 = \{p^1_1, p^1_2, ...p^1_k\}$, and define $L_2 = \{p^2_1, p^2_2, ...p^2_k\}$ similarly for another model $M_2$ on the same set of $k$ notes, where $p^i_j$ is a probability value in the range [0, 1]. From here, we can define lists of ranks, $R_1 = \{R^1_1, \ldots, R^1_k\}$ and $R_2 = \{R^2_1, \ldots, R^2_k\}$ where the elements represent the rank of each element in the list. Following the definition of Kendall (1945) and its implementation by SciPy (Virtanen et al., 2020), the Kendall's Tau $\tau_{(1,2)}$ between $L_1$ and $L_2$ is defined as:

$$\tau_{(1,2)} = \frac{P - Q}{\sqrt{(N - T)(N - U)}}$$

where $N = k(k-1)/2$ is the total number of pairs. $P$ represents the count of concordant pairs and $Q$ denotes the count of discordant pairs. $T$ is the number of tied pairs in $R_1$, and $U$ is the number of tied pairs in $R_2$. Here concordant pairs are all $(i, j)$ such that $R^1_i < R^1_j$ and $R^2_i < R^2_j$ or $R^1_i > R^1_j$ and $R^2_i > R^2_j$ while discordant pairs are all $(i, j)$ such that $R^1_i < R^1_j$ and $R^2_i > R^2_j$ or $R^1_i > R^1_j$ and $R^2_i < R^2_j$. $\tau$ is in the range $[-1, 1]$, where 1 indicates complete concordance, and -1 indicates complete discordance. We define Kendall's $\tau$-based distance as $d_{i,j} = (1 - \tau_{i,j})/2$. Since $\tau \in [-1, 1]$, it follows that $d \in [0, 1]$.

#### 3.2.2 Dataset-level information difference score

We perform multiple experiment runs per type of summarization and per hospital sites. The mean distance between the predictions of one summary type (intra-aspect difference) is calculated by averaging the distances of all combinations within the same settings. For example, the difference $D$ within $S_{plain}$ type summary experiments for a hospital $A$ can be defined as:

$$\mathrm{D}(A, S_{plain}) = \frac{\sum_{i=1}^{n} \sum_{j=1}^{n} d_{i,j}}{n^2}$$

where $n$ is the number of repeated training runs for the same settings with different random seeds (i.e. models $M_1, M_2, ...M_n$ are trained on $S_{plain}$).

[2]Note that these predictions are based on one experimental run, rather than the aggregation of multiple runs reported in Table 2.

[3]We utilized implementation of **SciPy v1.14.1**.

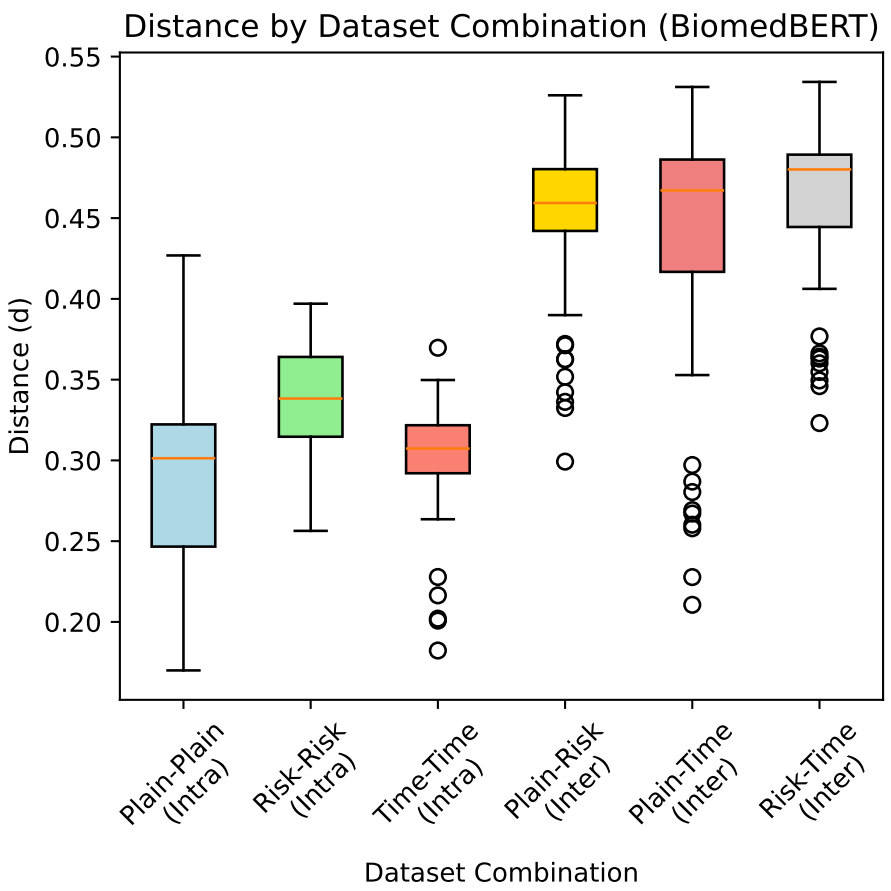

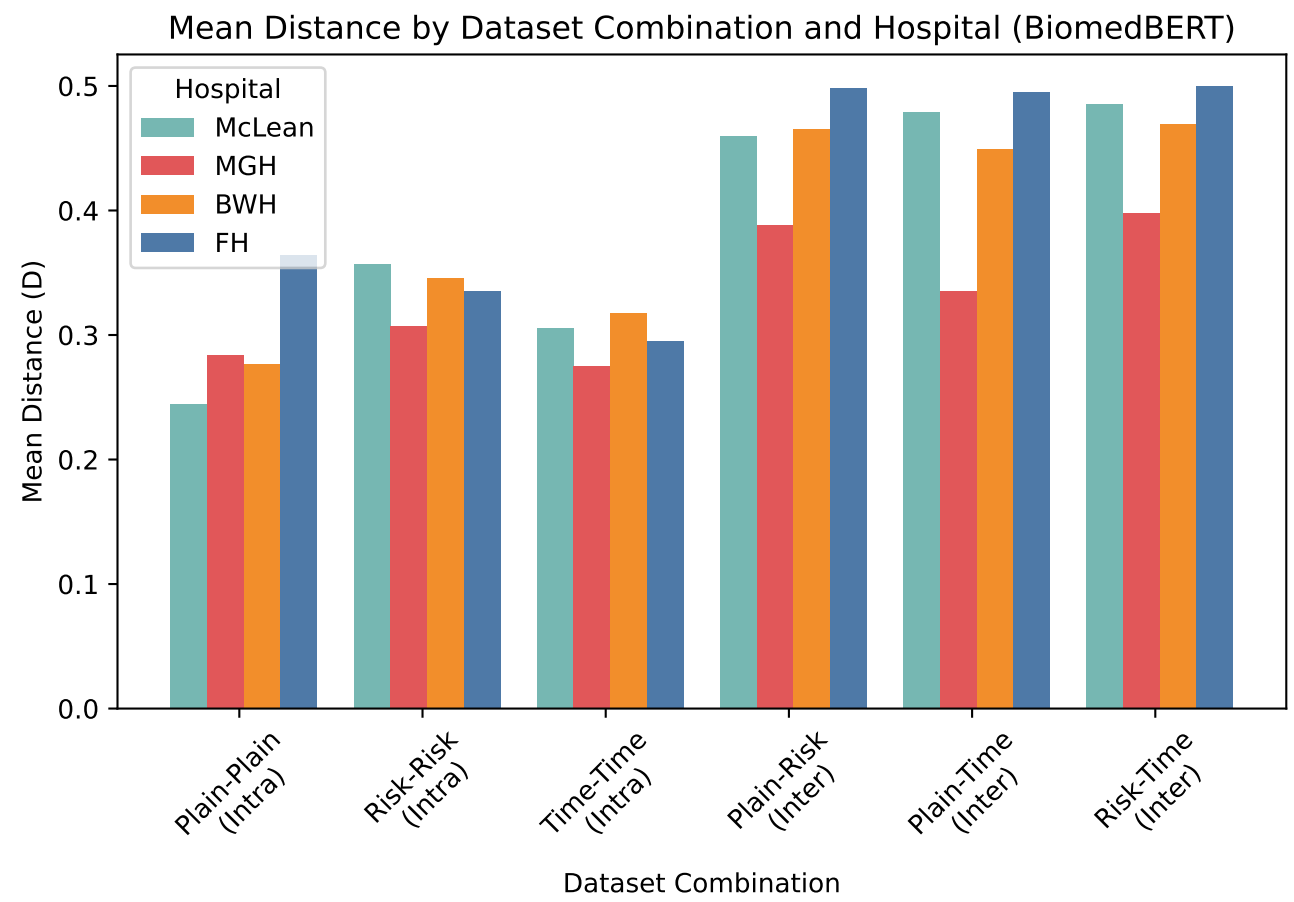

Figure 4: Information difference by summary type combinations. Lower values indicate smaller difference observed. [Left figure] The y-axis represents the Tau-based distance $d$. The box indicates the interquartile range (the middle 50% of the data), and the dots indicate outliers from the iterated experiments. [Right figure] The bar chart represents the Difference ($D$) (i.e. mean distance at the dataset level) by the summary type combination.

The mean distance between two different types of summarization methods (inter-aspect difference) is measured analogously. For example, the difference between Plain summarization and Riskfactor summarization can be measured by:

$$\mathrm{D}(A, S_{plain}, S_{riskfactor}) = \frac{\sum_{i=1}^{n} \sum_{j=1}^{m} d_{i,j}}{nm}$$

where $n$ is the number of repeated training runs for the Plain summarization dataset and $m$ for the Riskfactor dataset.

### 3.3 Readmission prediction task

In this section, we propose methods to utilizing the different types of summarized text for training task-specific models.

#### 3.3.1 Baseline models

Discharge notes are typically long usually requiring a 10k-token window, which exceeds the capacity of most encoder-only transformer models thus rendering fine-tuning impractical. Therefore, a Bag-of-Words (BoW) approach with Support Vector Machine (SVM) (Cortes, 1995; Joachims, 1998) was used as a baseline model to estimate the performance using the content of the entire document (i.e. no summarization).

Additionally, we experimented with supervised fine-tuning (using LoRA adapters (Hu et al., 2022)) of two relatively small LLMs capable of handling longer inputs. However, due to the nature of the training dataset, specifically, the number of training examples not large enough to train LLMs and the considerable length of each input, the fine-tuning process was challenging. The resulting model performance was unstable and often failed to yield meaningful results. (See Appendix C for details.) These outcomes further support our approach of summarizing discharge notes before training, as a way to overcome the limitations posed by full-document modeling in low-resource settings.

For baseline classification from single summaries, we used two types of models: (1) a traditional baseline with SVM with BoW features, and (2) a transformer-based Pretrained Language Model (PLM). The SVM baseline allows us to compare directly to full-document performance, while the PLM allows us to measure the capability of more advanced models.

#### 3.3.2 Integration of information from different summaries

We examine two methods of data integration: ***instance concatenation***, which we refer to as ***merged***, where we combine summaries at the note level; and ***dataset concatenation***, which we refer to as ***union***, where we create one large dataset where each instance has its own summary type.

Let $s_i^{plain}$ denote a plain summarized note at the $i$-th index, and likewise $s_i^{riskfactor}$ and $s_i^{timeline}$ for their respective summary types. The three datasets of the LLM-summarized notes, $S_{plain}$, $S_{riskfactor}$, and $S_{timeline}$, each contain $k$ instances that can be aligned.

The ***instance concatenation (merged)*** approach is represented as building a dataset composed of $s_i^{merged}$, which is three summaries, $s_i^{plain}$, $s_i^{riskfactor}$, $s_i^{timeline}$, simply concatenated together with summary demarcation indicated by "Another summary" to form one longer document.

$$s_i^{merged} = s_i^{plain} + s_i^{riskfactor} + s_i^{timeline}$$
$$S_{merged} = \{ s_i^{merged} \mid i = 1, 2, \dots, k \}$$

The ***dataset concatenation (union)*** approach is more straightforward and can be represented as the union of three summary datasets. This approach requires a pooling strategy during testing to combine the outputs of the summaries with the same original index and produce a single prediction for each original instance.

$$\begin{aligned} S_{union} &= S_{plain} \cup S_{riskfactor} \cup S_{timeline} \\ &= \{ s_1^{plain}, s_2^{plain}, \dots, s_k^{plain}, \\ &\qquad s_1^{riskfactor}, \dots, s_k^{timeline} \} \end{aligned}$$

We examined two prediction strategies for the dataset concatenation approach, ***Soft voting*** and ***Any voting***. A prediction for the $i$-th input can be expressed as follows, if the prediction and probability of the model for a summary type $s_i^{type}$ are given as $\hat{y}_i^{type} \in \{0, 1\}$ and $p_i^{type} \in [0, 1]$:

$$p_i^{softvote} = \text{average}(p_i^{plain}, p_i^{riskfactor}, p_i^{timeline})$$
$$\hat{y}_i^{softvote} = \begin{cases} 1, & \text{if } p_i^{\text{soft vote}} \geq 0.5, \\ 0, & \text{if } p_i^{\text{soft vote}} < 0.5 \end{cases}$$
$$p_i^{anyvote} = \max(p_i^{plain}, p_i^{riskfactor}, p_i^{timeline})$$
$$\hat{y}_i^{anyvote} = \hat{y}_i^{plain} \vee \hat{y}_i^{riskfactor} \vee \hat{y}_i^{timeline}$$

The number of elements in $S_{union}$ will be $3 * k$ whereas for $S_{merged}$ it will remain $k$. However, the length of each element for $S_{merged}$ will be longer than $S_{union}$ .

# 4 Results

## 4.1 Information differences

We conducted our experiments with encoder-only models as they produce a list of outputs with likelihood. Figure 4 shows results for BiomedBERT (formerly known as PubMedBERT) (Gu et al., 2020) which we fine-tuned for the specific task of readmission prediction. Aspect-oriented summarization was done with prompts to mistral 7B instruct v0.2 (Jiang et al., 2023) (see Appendix D for details).

Intra-aspect information difference scores (control group) were lower than inter-aspect difference scores (experimental group). This implies that the different aspect-oriented prompts produce summaries with non-overlapping information, and this is not the effect of randomness introduced during initialization steps. Further, these results provide a strong justification for developing methods that integrate signals from aspect-oriented summaries.

In addition to experimenting with BiomedBERT, we experimented with Clinical Longformer (CLF) (Li et al., 2023). A detailed description is provided in Appendix E. The result patterns for CLF are consistent with those of BiomedBERT.

## 4.2 Readmission prediction

**Evaluation metrics** We utilized multiple evaluation metrics, including threshold-dependent metrics (precision, recall, and F1-score) and threshold-independent metrics such as area under the receiver operating characteristic (AUROC) and area under the precision-recall curve (AUPRC). For the threshold-dependent metrics, we used 0.5 as a fixed threshold for positive and negative labeling. We prioritized threshold-independent metrics as our primary evaluation criteria, as threshold-dependent metrics e.g. F1-score are highly sensitive to the choice of threshold in the presence of class imbalance—a characteristic of all our datasets (see Table 1). We examined threshold-dependent metrics mainly to understand how the model performs at the default threshold. We also report F1 scores for the negative class (denoted as Neg F1) and Macro average of positive and negative F1 scores (denoted as MaF1) as an alternative way to address the limitations of the F1-score.

Table 2 shows the results of models trained on aspect-oriented summaries and our proposed data integration approaches for the 30-day readmission prediction task. The columns are grouped into two types of supervised models, SVM with BoW features and fine-tuned BiomedBERT Transformer models. Regardless of model types, models trained on the integrated dataset showed strong performance over those trained on monotone summarization datasets, where summaries were generated

| Model types | | SVM with BoW | | | | | Transformer Model (BiomedBERT) | | | | |
|---|---|---|---|---|---|---|---|---|---|---|---|
| Input type | | Full note | Plain | Riskfactor | Timeline | *Merged* | Plain | Riskfactor | Timeline | *Soft vote* | *Any vote* |
| McLean | AUROC | 0.5770 | **0.5784** | 0.5415 | 0.5668 | **0.5853** | **0.6051** | 0.5193 | 0.5788 | **0.6005** | 0.5988 |
| | AUPRC | 0.3465 | **0.3541** | 0.3164 | 0.3490 | **0.3717** | **0.3960** | 0.3074 | 0.3650 | 0.3868 | **0.3943** |
| | MaAvg F1 | 0.5570 | 0.5531 | 0.5346 | **0.5704** | **0.5728** | **0.5630** | 0.5113 | 0.5500 | **0.5694** | 0.5122 |
| | *Neg F1* | 0.7450 | 0.7390 | 0.7313 | 0.7392 | 0.7459 | 0.7695 | 0.7617 | 0.7748 | 0.7721 | 0.5887 |
| | *Pos F1* | 0.3689 | 0.3671 | 0.3379 | 0.4016 | 0.3997 | 0.3566 | 0.2608 | 0.3252 | 0.3667 | 0.4357 |
| MGH | AUROC | 0.5923 | 0.6075 | 0.5565 | **0.6102** | **0.6233** | 0.6476 | 0.5890 | 0.6494 | **0.6672** | **0.6564** |
| | AUPRC | **0.4208** | 0.4123 | 0.3447 | 0.4127 | **0.4233** | 0.4730 | 0.3988 | 0.4809 | **0.4870** | **0.4823** |
| | MaAvg F1 | **0.6120** | **0.6022** | 0.5462 | 0.5949 | 0.5965 | 0.6020 | 0.5477 | 0.6006 | **0.6025** | **0.6025** |
| | *Neg F1* | 0.7491 | 0.7330 | 0.7153 | 0.7351 | 0.7401 | 0.7684 | 0.7622 | 0.7746 | 0.8044 | 0.7144 |
| | *Pos F1* | 0.4749 | 0.4715 | 0.3772 | 0.4547 | 0.4529 | 0.4355 | 0.3333 | 0.4267 | 0.4007 | 0.4907 |
| BWH | AUROC | 0.5420 | **0.5601** | 0.5312 | **0.5527** | 0.5518 | 0.5402 | 0.5174 | 0.5677 | **0.5938** | **0.5966** |
| | AUPRC | 0.3803 | **0.4014** | 0.3590 | **0.4032** | 0.3939 | 0.3694 | 0.3457 | 0.3856 | **0.4124** | **0.4284** |
| | MaAvg F1 | 0.5520 | 0.5485 | 0.5196 | **0.5674** | **0.5669** | 0.5383 | 0.5052 | **0.5404** | **0.5518** | 0.5206 |
| | *Neg F1* | 0.6982 | 0.6855 | 0.6779 | 0.6944 | 0.6941 | 0.6901 | 0.6678 | 0.7168 | 0.7295 | 0.5474 |
| | *Pos F1* | 0.4059 | 0.4116 | 0.3614 | 0.4404 | 0.4396 | 0.3865 | 0.3426 | 0.3640 | 0.3741 | 0.4938 |
| FH | AUROC | **0.5750** | 0.5413 | 0.5115 | 0.5290 | **0.5646** | 0.4800 | 0.4905 | **0.5350** | **0.5353** | 0.5336 |
| | AUPRC | **0.4190** | 0.3473 | 0.3224 | **0.3541** | 0.3497 | 0.3007 | 0.3138 | **0.3549** | 0.3415 | **0.3528** |
| | MaAvg F1 | 0.5703 | **0.6041** | 0.4935 | **0.6280** | 0.4861 | 0.4558 | **0.4755** | 0.4671 | 0.4451 | **0.5126** |
| | *Neg F1* | 0.7182 | 0.6899 | 0.6849 | 0.6909 | 0.6860 | 0.7426 | 0.7106 | 0.7721 | 0.8027 | 0.6996 |
| | *Pos F1* | 0.4223 | 0.5183 | 0.3022 | 0.5652 | 0.2862 | 0.1691 | 0.2404 | 0.1620 | 0.0874 | 0.3257 |

Table 2: Performance of supervised models by summarization methods. **Boldfaced** numbers indicate the top-2 performances for each model type, while **underlined** numbers denote the best performances. Columns titled Merged, Soft vote, and Any vote show the performance of models using three types of summarized inputs. Pos F1 refers to the F1-score for the positive label (patient readmitted within 30 days), while Neg F1 represents the F1-score for the negative label (patient NOT readmitted within 30 days). MaAvg F1 denotes the macro-average of Neg F1 and Pos F1.

using a single-type aspect-oriented prompt.

The first column presents the results of the baseline approach (SVM with BoW features) using full discharge notes. These results, with AUROC values ranging from 0.54 to 0.59, indicate that the task is highly challenging, yet there are detectable signals that the models can learn.

For the SVM with BoW features results on summarized discharge notes, the merging method showed a performance gain across all datasets. For Transformer models, soft voting and any voting showed a large improvement over the same architecture models trained on three summary types.

The training runs were performed 5 times (BiomedBERT) or 10 times (SVM with BoW) using different random seeds while keeping the hyperparameters identical to ensure statistical robustness against the effects of random initialization. All reported numbers represent the averaged values of these runs. Additional statistics for these results are provided in Table 4 in Appendix F.

## 5 Discussion

### 5.1 Integration method by model types

We reported different integration methods depending on the model type: instance concatenation (Merged) for SVM with BoW, and dataset concatenation (Soft voting and Any voting) for transformer-based models. Because the Merged method increases input sequence length—sometimes up to 1,500 tokens (with a median range of 870 to 950 across hospitals)—we applied it only to Clinical Longformer (Li et al., 2023), an encoder-only transformer that can handle inputs longer than 512 tokens. The results of this experiment are provided in Appendix G (Table 5).

Our findings show that the Merged approach improves performance for the BoW model but does not benefit transformer models. One likely reason is that BoW methods are relatively unaffected by input length, whereas transformer models often exhibit degraded performance on lengthy inputs (Beltagy et al., 2020; Ainslie et al., 2020; Li et al., 2023; Yoon et al., 2024). When tested on Clinical Longformer, the Merged method was less effective than the Union method. Among Soft Voting and Any Voting, the results followed a similar pattern to our main findings with BiomedBERT (Table 2).

### 5.2 Zero-shot LLM prompting

An alternative to our classification pipeline is using LLMs to directly predict readmission. We tested the ability of an LLM, `Llama-3.1-8B-Instruct` (Dubey et al., 2024), and report its performance on zero-shot short-term readmission prediction in Table 6 in Appendix H.

The zero-shot results do not show superior performance when compared with supervised models. This is reasonable for the zero-shot prompting scenario as the LLM is unlikely to have been exposed to similar data during pre-training and thus cannot learn the distribution without labels. To support this argument, we examined the proportion of positive labels across LLM predictions, supervised model predictions, and gold-standard datasets, following Yoon et al., 2024. (See Appendix H, Figure 8, for the results plot.) Compared to the Transformer model predictions, the proportion of positive labels is notably misaligned in the LLM model predictions. This supports our assumptions that one of the reasons LLMs exhibit sub-optimal results is due to their misalignment with the true label distribution.

### 5.3 Performance of supervised models on raw discharge notes

At the first stage of our study, we examine the performance of supervised models trained directly on raw discharge notes (i.e., without summarization) across hospitals. These results serve as a baseline and provide preliminary evidence for the benefits of incorporating summarization prior to classification.

We experimented with SVM with BoW, a hierarchical transformer (HT) model (Su et al., 2021), and a Clinical-Longformer (CL)-based approach. The performance of the SVM and HT models is reported in Table 7 in Appendix I. In our experiments, the CL approach trained on our dataset with the first 4k tokens was not successful, likely because the length of our instances substantially exceeded the token window. As shown in the table, the HT model did not outperform the SVM with BoW baseline. For this reason, and due to space constraints, we included the SVM with BoW as the representative raw note baseline in the main text.

### 5.4 Hallucination risks and mitigation

While our use of LLMs is limited to summarization conditioned on discharge notes, hallucinations remain a critical concern. Large language models (LLMs) may produce content that is not fully grounded in the input text. In the medical domain, such hallucinations are particularly concerning because inaccurate or fabricated information can propagate into high-stakes downstream tasks and negatively affect clinical decision-making (Kim et al., 2025). Future work toward clinical applications should therefore investigate approaches to verify factual consistency and develop complementary strategies to reduce hallucination risks.

## 6 Related work

### 6.1 Prompting strategies

With the recent advance of LLMs, a growing body of research has focused on designing effective prompting strategies to better leverage their capabilities. A foundational work by Lester et al., 2021 introduced *prompt tuning*, a method that learns soft prompts while keeping the model parameters fixed. Since then, one research stream has aimed toward dynamic or adaptive prompting, a term proposed in contrast to static, hard-coded prompts. For example, Yang et al., 2023 and Nehring et al., 2024 propose frameworks that adapt prompts based on input context or task complexity.

Another direction is self-consistency prompting, which improves reasoning performance by generating multiple reasoning paths and selecting the most consistent answer. A significant work by Wang et al., 2022 applied self-consistency to Chain-of-Thought (CoT) prompting, and more recently, Chen et al., 2023 proposed Universal Self-Consistency, extending the method to general generation tasks beyond CoT.

Closely related to prompt tuning is the task of Aspect-Oriented Summarization (AOS). This task can be traced back to the work of Hu and Liu, 2004, which introduced feature-focused summarization for customer reviews. More recently, Ahuja et al., 2022 proposed ASPECTNEWS, a modern AOS dataset and benchmark that aligns closely with our work, particularly in settings such as single-document summarization with multiple targeted aspects. This was followed by datasets such as ACLSum (Takeshita et al., 2024), which further expanded the scope of aspect-aware summarization to scientific publications.

While methods like dynamic prompting aim to optimize prompt selection, identifying and combining multiple prompts that capture diverse and complementary information signals remains an open challenge. In this work, we address this gap by proposing an automatic framework for quantifying informational differences between prompts, enabling more effective integration of multi-perspective summaries for downstream tasks.

### 6.2 Clinical text summarization

Our research lies at the intersection of clinical NLP and summarization. Work on clinical text summarization has a long history, with foundational research addressing the complexities of healthcare data. One example is McKeown et al. 1997, where the authors developed systems to generate multimedia summaries to support time-pressured caregivers. Subsequent research introduced natural language generation techniques for summarizing time-series clinical text data to assist in neonatal care (Sripada et al., 2003; Portet et al., 2009). These early systems addressed key challenges in content selection. A comprehensive overview by Pivovarov and Elhadad 2015 provides an important summary of EHR summarization methods, capturing the state of the field prior to the emergence of LLMs.

Recent studies have demonstrated the strong capabilities of LLMs in clinical summarization tasks. For instance, Van Veen et al., 2024 demonstrated that summaries generated by LLMs can be more complete and concise than those written by clinicians. Similarly, Ellershaw et al., 2024 proposed a method for generating discharge summaries by guiding LLMs with structured clinical guidelines, highlighting the potential of combining domain knowledge with generative models.

### 6.3 Patient outcome prediction in Clinical NLP

Patient outcome prediction is a core task in clinical NLP, encompassing subproblems such as mortality prediction and other outcome measures. Our work focuses specifically on readmission prediction. In this area, the ClinicalBERT paper (Huang et al., 2019) proposed fine-tuning of transformer model to various types of clinical notes to predict readmission. While ClinicalBERT focused on relatively short notes as input, LCD Benchmark paper (Yoon et al., 2024) introduced another challenging task: predicting out-of-hospital mortality from long-form discharge notes. This study did not incorporate summarization; however, it highlighted the challenges of outcome prediction with lengthy documents, which motivates our use of summarization in this context.

## 7 Conclusion

In this study, we explored a method for processing long documents using aspect-oriented summarization aimed to capture different views of the information of the original document. Our study provides three key insights: (1) we hypothesize that LLM summaries generated with different aspect-oriented prompts hold different details or *information signals*, (2) we propose methods to measure these signals, (3) we investigate methods to effectively integrate signals from different types of summaries for supervised training of transformer models. We applied our methodology to the high-impact task of 30-day psychiatric re-admission prediction.

## Limitations

We collected datasets from four different sites in the US. Two of the most notable differences between those datasets are dataset size and the positive-label ratio. However, the datasets differ in the length of the notes. In this study, we did not account for these variations and leave it as future work.

This study utilized only one type of EHR documentation – discharge notes. Other types of notes (e.g., progress notes) and additional data modalities (e.g., structured data) are available but not included in this paper. In terms of prompt diversity, our aspect-oriented summarization prompts could be further enriched with other critical aspects for readmission prediction, e.g. medications, trauma history and comorbidities.

Our current study is a research into the important topic of 30-day psychiatric readmission prediction. It is not an application ready for direct clinical applications. Such clinical applications require carefully designed clinical trials involving multiple domain experts. In the medical domain, hallucinations carry particular risks because misleading or fabricated information may directly affect patient safety and clinical decision-making. Since our approach uses LLMs to summarize input text, it is not entirely free from the risk of hallucinations.

## Acknowledgments

Research reported in this publication was supported by the National Institute Of Mental Health and National Library of Medicine of the National Institutes of Health under Award Numbers R01MH126977 and R01LM012973. The content is solely the responsibility of the authors and does not necessarily represent the official views of the National Institutes of Health. This work was partially supported by Korean Institute for Advancement of Technology (KIAT) grant RS-2024-00435997. Chanwhi

Kim participated in this work while he was visiting MLML Lab, led by Tim Miller at Boston Children's Hospital and Harvard Medical School.

# Appendix

## A Prompt template

One of the prompt templates, riskfactor-focused prompt, is shown in Figure 5. Other prompt templates are shown in Figure 2 of main text Section 3.1.

## B Sample psychiatric note

Figure 6 shows three sample summaries generated from a psychiatric note using our specified prompt. All numerical values and dates presented in this appendix have been replaced with randomized placeholders for illustrative purposes. Due to this process, some dates (e.g., those in the year 2030) may fall in the future relative to the time of submission.

## C Baseline - Supervised fine-tuning of LLMs using LoRA adapters

As an alternative baseline in this study, we explored direct supervised fine-tuning of relatively small-scale LLMs. This experiment was designed to test the viability of full-document modeling and to support our premise that summarization may be a more effective strategy for discharge note classification.

We fine-tuned the models using Low-Rank Adaptation (LoRA) adapters (Hu et al., 2022). Specifically, we used two instruction-tuned models capable of handling extended context: Qwen-2.5-0.5B-Instruct (Qwen et al., 2024) and Qwen3-0.6B (Yang et al., 2025). The LoRA configuration used a rank of 12 and targeted the `["q_proj", "v_proj"]` modules.

The models were fine-tuned for a binary sequence classification task: predicting whether a patient would be readmitted within 30 days (label 1) or not (label 0). Training was conducted for 10 epochs using the full discharge notes as input.

Table 3 presents the results in terms of AUROC and Positive-class F1 score (PosF1) for each hospital dataset. The results reported reflect a single run for each model. Despite their larger size and the ability to process full discharge notes thanks to extended input length capacity, both models demonstrated limited and unstable performance across all sites and evaluation metrics. Performance varied substantially across institutions and was often close to random chance, with AUROC values near or below 0.5 on the test set. Notably, AUROC values below 0.5 suggest that the model may have failed to learn meaningful patterns during training. However, it is worth noting that AUROC values on the development set were mostly above 0.5, indicating a potential mismatch between the training distribution and the unseen test data, or overfitting due to the limited dataset size. F1 scores on positive labels (*Pos F1*) were particularly low, highlighting the models' difficulty in accurately identifying positive cases.

These findings suggest that, under the current training setup and dataset constraints, supervised fine-tuning of small LLMs using LoRA does not serve as a reliable baseline, and further supporting our approach of using summarization followed by model training. Potential contributing factors include class imbalance, limited dataset size, and the models' restricted capacity to generalize within the complex, domain-specific language of clinical documentation.

## D Experimental details

**Summarization:** Full discharge notes are summarized using a quantized version of mistral 7b instruct v0.2 [4] with `llama.cpp` library. Inference speed varies, but even for the largest dataset, processing was completed within a day on a workstation with a single NVIDIA GeForce RTX 4090 GPU. The generated summaries were always fewer than 512 tokens, and the median number of tokens varied by hospital and summary type, though all fell within the range of 256 to 356.

**Supervised training:** For SVM with BoW features, we used the scikit-learn library (Pedregosa et al., 2011). For transformer models, we employed the Clinical NLP Transformers library v0.7.0 (CNLPT, 2021)[5], which wraps the Hugging Face Transformers library for training and testing (Wolf et al., 2019). For pretrained weights, we utilized the abstracts version of BiomedBERT (Gu et al., 2020).[6]

### D.1 Transformer model hyperparameter settings

For transformer models, we performed a minimal hyperparameter search using the development

[4] `mistral-7b-instruct-v0.2.Q4_K_M.gguf`

[5] Available at: https://github.com/Machine-Learning-for-Medical-Language/cnlp_transformers

[6] `microsoft/BiomedNLP-BiomedBERT-base-uncased-abstract-fulltext`

**Riskfactor-focused summarization prompt**

<s>[INST] The following is a discharge summary for a patient leaving a psychiatric hospital.
***Discharge note inserted here.***
This is the end of the discharge summary. We would like to summarize the discharge summary in terms of risk factors for readmission within 30 days. Psychiatric experts have created a list of risk factors that they believe are important.
1: Appearance: Physical appearance, gestures, and mannerisms;
2: Thought content: Suicidal/homicidal ideation, obsessions, phobias, delusions, hallucinations;
3: Interpersonal: Family situation, friendships, and other social relationships;
4: Mood: Feelings and overall disposition;
5: Occupation: School and/or employment;
6: Thought process: Pace and coherence of thoughts. Includes linear, goal-directed, perseverative, tangential, and flight of ideas;
7: Substance: Drug and/or alcohol use;
8: Other: Any risk factor that does not fall into one of the preceding categories;
These risk factors could have positive or negative valence. Please summarize the risk factors from the above list that are present in the discharge summary, along with their valence (positive, negative, neutral). [/INST]

Figure 5: Prompts for summarization. Three types of prompts, `plain`, `riskfactor`, and `timeline`, were used to generate different types of summaries.

**Plain-prompt summary**

The patient, a **-year-old individual, was admitted to the psychiatric hospital for a major depressive episode with suicidal ideation. Upon admission, they were assessed to have moderate depression severity with significant impairment in daily functioning. The patient received a combination of pharmacological treatment and psychotherapy during their stay. They were initially prescribed **** and later added ***** due to persistent symptoms. Psychotherapy sessions focused on cognitive behavioral techniques and interpersonal skills training. The patient showed gradual improvement and was able to engage in activities of daily living and maintain a stable mood by the time of discharge. They were advised to continue outpatient treatment and follow-up appointments with their mental health care provider.

**Riskfactor-prompt summary**

Based on the provided discharge summary, the following risk factors and their valence are identified:
1. Thought content: The patient reports passive suicidal thoughts that are baseline for her, but no intent or plan; this can be considered a negative risk factor.
2. Substance: The patient has a history of alcohol use and relapse during the hospitalization, which is a negative risk factor.
3. Occupation: The patient's academic progress and employment status were not mentioned in the summary, so this risk factor is neutral.
4. Interpersonal: The patient has supportive family and close follow-up with established treaters, which are positive risk factors.
5. Mood and overall disposition: The patient expresses anxiety and feels ready for discharge, which can be considered neutral risk factors.
6. Thought process: No abnormalities were reported in the patient's thought process during the hospitalization, so this risk factor is neutral.
7. Other: The patient's close proximity to a liquor store and chronic suicidal thoughts are negative risk factors.
8. Appearance, gestures, and mannerisms: No information about the patient's appearance was provided, so this risk factor is neutral.

**Timeline-prompt summary**

* 12/31/19**: Born
* Unknown date: Diagnosed with Schizoaffective disorder, bipolar type
* Unknown date: Diagnosed with borderline personality disorder
* Unknown date: Diagnosed with polysubstance disorder, including alcohol and cannabis use
* Unknown date: Multiple prior hospitalizations
* Unknown date: History of self-injury by cutting, no known suicide attempts
* 9/12/2030: Admitted to psychiatric hospital for evaluation and treatment of confusion/disorganization in the context of worsening depression and alcohol/cannabis use
* 9/28/2030: Discharged from psychiatric hospital, diagnosed with Schizoaffective disorder, bipolar type, and prescribed lithium carbonate *** mg capsule nightly at bedtime.

Figure 6: Summarized notes. All numerical values and dates presented in this appendix have been replaced with randomized placeholders. Each summaries are generated using three types of prompts, `plain`, `riskfactor`, and `timeline`

| Dataset | Metrics | Qwen-2.5-0.5B-Instruct Devel. | Test | Qwen3-0.6B Devel. | Test |
|---|---|---|---|---|---|
| McLean | AUROC | 0.5203 | 0.4887 | 0.5692 | 0.5452 |
| | Pos F1 | 0.0513 | 0.1184 | 0.2362 | 0.2298 |
| MGH | AUROC | 0.6549 | 0.6109 | 0.5878 | 0.5584 |
| | Pos F1 | 0.2032 | 0.1098 | 0.1327 | 0.0963 |
| BWH | AUROC | 0.3986 | 0.4893 | 0.4908 | 0.5595 |
| | Pos F1 | 0.3385 | 0.2275 | 0.4058 | 0.4369 |
| FH | AUROC | 0.6461 | 0.4480 | 0.5271 | 0.3883 |
| | Pos F1 | 0.2286 | 0.0430 | 0.2222 | 0.0238 |

Table 3: Performance of supervised fine-tuned Qwen-2.5-0.5B-Instruct and Qwen3-0.6B (instruction tuned) using LoRA adapters. The columns labeled Devel. and Test show model performance on the development and test datasets, respectively.

dataset for the smaller datasets BWH and FH Specifically, we explored the following hyperparameters:

- Learning Rate: {2e-5, 5e-6, 2e-6}; Epochs: {40, 200}.

For the final reported performance, we applied the same hyperparameter settings across all datasets and pretrained language models:

- Learning Rate: 2e-5; Epochs: 40; Batch Size: 8.

For the iterated experiments, random seeds were incrementally set starting from 0. For example, in five iterations for BiomedBERT, the seeds were {0, 1, 2, 3, 4}. During training, the model was evaluated on the development set using loss, and the best checkpoint was saved. All results reported in this paper are based on the test dataset split.

## E Information difference for Clinical Longformer

This section provides details on our additional experiments conducted using the Clinical Longformer (Li et al., 2023). We conducted 10 iterations of experiments for models initialized with the Clinical Longformer. Figure 7 shows the information difference scores for Clinical Longformer model outputs. The results are consistent with those provided in the main text, Section 4.1.

## F Statistics of iterated experiments

The transformer model results in the Table 2 show averaged scores of 5 runs, and their standard deviation is in Table 4.

## G Supplementary results for Section 5.1

In this section we present Table 5 to support the discussions in Section 5.1.

## H Zero-shot LLM prompting results

Table 6 reports the performance of zero-shot LLM-based prediction of readmission likelihood, using verbal (i.e., string-formatted) outputs. Main text of this section is in Section 5.2

## I Performance of supervised models on raw discharge notes

In this section, we present experiments with supervised models trained on raw discharge notes (i.e. notes that are not summarized) across hospitals in Table 7.

## J Comparison of short-term and long-term patient outcome prediction

In general, the length of the follow-up period until an outcome event occurs is an important factor that affects both the characteristics and the difficulty of patient outcome prediction tasks. Which factors are more influential in short-term versus long-term prediction depends on the specific task, and is outside the scope of this study. However, short-term prediction tasks often show different characteristics compared to long-term prediction tasks because of differences in prevalence. For example, in the Pos/Total column of Table 1, the proportion of positive cases in our task is about 22–34% of admissions, which is comparable to prior findings—for instance, Pederson et al. 2016 reported a 30-day readmission rate of 16.2% (20.4% for patients with depressive symptoms)—whereas in long-term or lifetime readmission prediction the prevalence is substantially higher (Burcusa and Iacono, 2007). Another difference lies in the confounding factors, such as deaths due to natural causes in elderly patients, or acute deterioration and clinical uncertainty. Therefore, short-term and long-term patient outcome prediction tasks can be regarded as distinct problems.

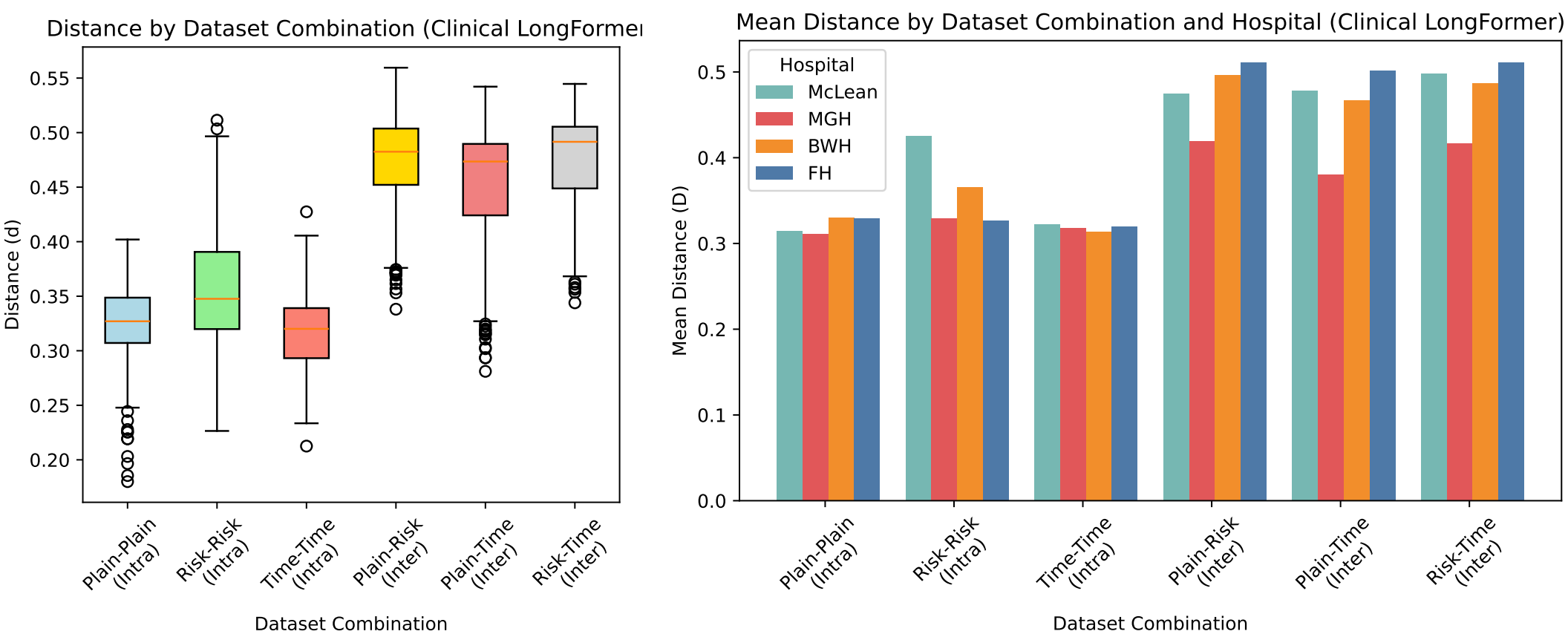

Figure 7: Information difference for Clinical Longformer.

| Hospital | Summ Type | AUROC | AUPRC | MaAg_F1 | *Neg_F1* | *F1* | Group Size |
|---|---|---|---|---|---|---|---|
| McLean | Plain | 0.0084 | 0.0162 | 0.0105 | 0.0225 | 0.0257 | 5 |
| | Riskfactors | 0.0101 | 0.0130 | 0.0097 | 0.0157 | 0.0296 | 5 |
| | Timeline | 0.0129 | 0.0156 | 0.0134 | 0.0384 | 0.0522 | 5 |
| | Soft voting | 0.0075 | 0.0150 | 0.0064 | 0.0133 | 0.0201 | 5 |
| MGH | Plain | 0.0265 | 0.0367 | 0.0150 | 0.0277 | 0.0344 | 5 |
| | Riskfactors | 0.0132 | 0.0142 | 0.0074 | 0.0238 | 0.0377 | 5 |
| | Timeline | 0.0371 | 0.0402 | 0.0213 | 0.0230 | 0.0500 | 5 |
| | Soft voting | 0.0164 | 0.0259 | 0.0217 | 0.0098 | 0.0446 | 5 |
| BWH | Plain | 0.0157 | 0.0147 | 0.0131 | 0.0286 | 0.0427 | 5 |
| | Riskfactors | 0.0200 | 0.0248 | 0.0360 | 0.0477 | 0.0764 | 5 |
| | Timeline | 0.0135 | 0.0118 | 0.0178 | 0.0251 | 0.0453 | 5 |
| | Soft voting | 0.0139 | 0.0094 | 0.0102 | 0.0308 | 0.0234 | 5 |
| FH | Plain | 0.0252 | 0.0195 | 0.0357 | 0.0595 | 0.1037 | 5 |
| | Riskfactors | 0.0413 | 0.0306 | 0.0173 | 0.0505 | 0.0484 | 5 |
| | Timeline | 0.0200 | 0.0222 | 0.0302 | 0.0198 | 0.0747 | 5 |
| | Soft voting | 0.0217 | 0.0074 | 0.0338 | 0.0066 | 0.0724 | 5 |

Table 4: Standard deviation of the performance of iterated supervised transformer model, corresponding to the averages shown in Table 2 of the main paper.

| Input type | | Plain | Risk | Time | ***Merge*** | ***Soft vote*** | ***Any vote*** |
|---|---|---|---|---|---|---|---|
| McLean | AUROC | 0.587 | 0.525 | 0.569 | 0.576 | **0.606** | **0.599** |
| | AUPRC | **0.389** | 0.309 | 0.343 | 0.387 | **0.386** | 0.381 |
| | Ma F1 | 0.556 | 0.510 | 0.532 | **0.560** | **0.564** | 0.533 |
| | *NegF1* | 0.778 | 0.752 | 0.786 | 0.782 | 0.776 | 0.644 |
| | *PosF1* | 0.335 | 0.268 | 0.278 | 0.338 | 0.352 | 0.422 |
| MGH | AUROC | 0.633 | 0.584 | 0.615 | 0.612 | **0.675** | **0.668** |
| | AUPRC | 0.440 | 0.372 | 0.417 | 0.417 | **0.492** | **0.489** |
| | Ma F1 | 0.595 | 0.556 | 0.583 | 0.585 | **0.601** | **0.617** |
| | *NegF1* | 0.791 | 0.758 | 0.767 | 0.779 | 0.814 | 0.764 |
| | *PosF1* | 0.400 | 0.354 | 0.399 | 0.392 | 0.389 | 0.470 |
| BWH | AUROC | 0.547 | 0.540 | **0.563** | 0.543 | **0.604** | 0.591 |
| | AUPRC | 0.373 | 0.372 | 0.395 | 0.369 | **0.419** | **0.406** |
| | Ma F1 | 0.530 | 0.514 | **0.537** | 0.527 | **0.549** | 0.528 |
| | *NegF1* | 0.711 | 0.675 | 0.706 | 0.710 | 0.732 | 0.572 |
| | *PosF1* | 0.349 | 0.353 | 0.368 | 0.343 | 0.366 | 0.483 |
| FH | AUROC | 0.480 | 0.526 | **0.526** | 0.483 | 0.507 | 0.506 |
| | AUPRC | 0.310 | 0.331 | **0.348** | 0.310 | **0.333** | 0.322 |
| | Ma F1 | 0.455 | **0.498** | 0.497 | 0.460 | 0.438 | **0.499** |
| | *NegF1* | 0.758 | 0.745 | 0.757 | 0.765 | 0.775 | 0.662 |
| | *PosF1* | 0.152 | 0.251 | 0.237 | 0.155 | 0.100 | 0.336 |

Table 5: Performance of the Clinical Longformer model evaluated using different summarization methods.

| Input type | | Full | Plain | Risk | Time | Merged |
|---|---|---|---|---|---|---|
| McLean | Ma F1 | 0.5650 | 0.5458 | 0.4087 | 0.3866 | 0.3983 |
| | *NegF1* | 0.7696 | 0.7166 | 0.3840 | 0.3338 | 0.3818 |
| | *PosF1* | 0.3604 | 0.3750 | 0.4335 | 0.4394 | 0.4147 |
| MGH | Ma F1 | 0.4708 | 0.4895 | 0.3548 | 0.3707 | 0.3252 |
| | *NegF1* | 0.5312 | 0.5532 | 0.2772 | 0.2927 | 0.2011 |
| | *PosF1* | 0.4104 | 0.4257 | 0.4325 | 0.4487 | 0.4493 |
| BWH | Ma F1 | 0.4794 | 0.4590 | 0.3892 | 0.3527 | 0.3706 |
| | *NegF1* | 0.5792 | 0.4845 | 0.2847 | 0.2420 | 0.2388 |
| | *PosF1* | 0.3796 | 0.4334 | 0.4937 | 0.4635 | 0.5024 |
| FH | Ma F1 | 0.5093 | 0.4565 | 0.3688 | 0.2604 | 0.2986 |
| | *NegF1* | 0.7465 | 0.5149 | 0.2462 | 0.0465 | 0.1304 |
| | *PosF1* | 0.2720 | 0.3981 | 0.4913 | 0.4744 | 0.4667 |

Table 6: Performance of zero-shot LLM prompting for readmission prediction. Note that we cannot calculate AUROC or AUPRC because our LLM evaluation setting only provides binary predictions and therefore, ranking of predictions is not possible.

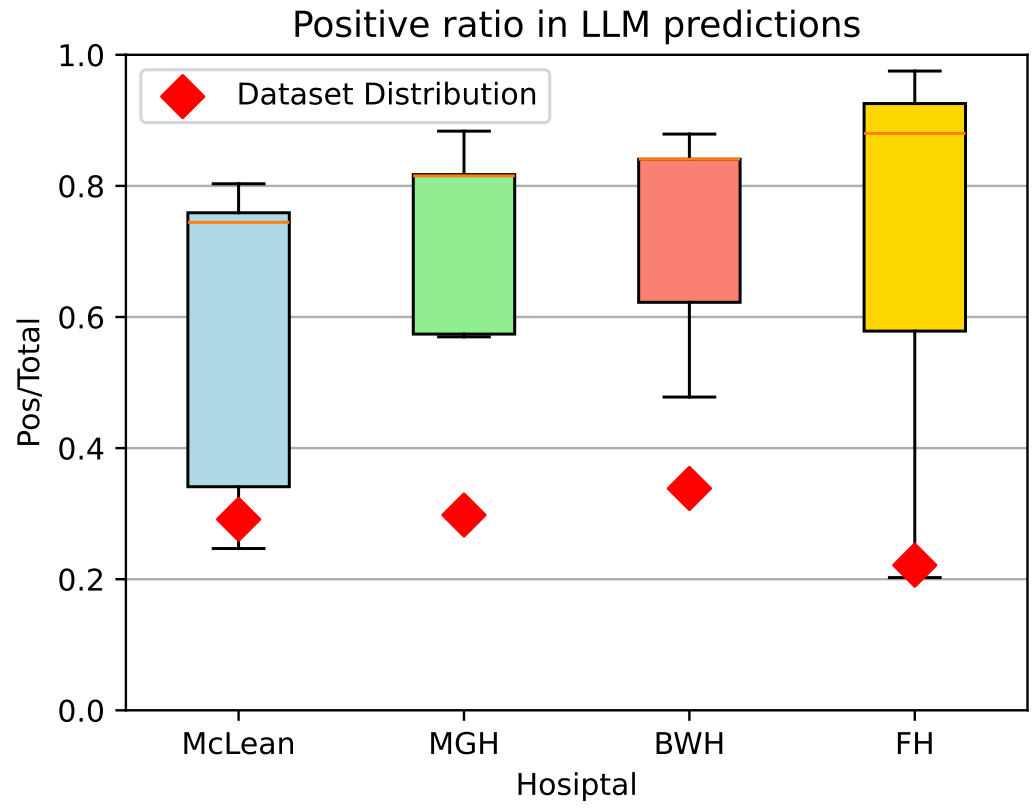

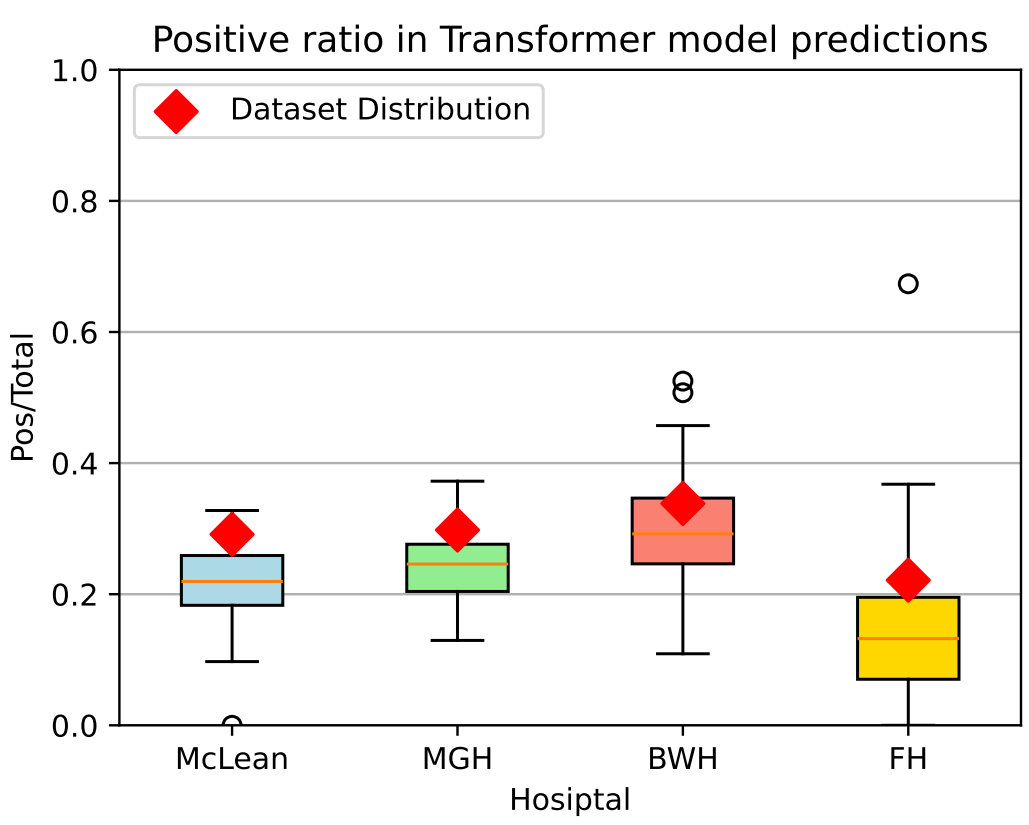

Figure 8: Boxplots showing the ratio of positive-label predictions in the model outputs. The red rhombus symbol indicates the ratio in the dataset. The box represents the interquartile range of the ratio values in the iterated experiments.

| Location | Metric | BoW+SVM | Hier-transformer |
|---|---|---|---|
| McLean | AUROC | 0.5770 | 0.5346 |
| | AUPRC | 0.3465 | 0.3136 |
| | MaAvg F1 | 0.5570 | 0.5224 |
| | *NegF1* | 0.7450 | 0.7581 |
| | *PosF1* | 0.3689 | 0.2867 |
| MGH | AUROC | 0.5923 | 0.5475 |
| | AUPRC | 0.4208 | 0.3435 |
| | MaAvg F1 | 0.6120 | 0.5343 |
| | *NegF1* | 0.7491 | 0.7596 |
| | *PosF1* | 0.4749 | 0.3090 |
| BWH | AUROC | 0.5420 | 0.5316 |
| | AUPRC | 0.3803 | 0.3579 |
| | MaAvg F1 | 0.5520 | 0.5043 |
| | *NegF1* | 0.6982 | 0.7285 |
| | *PosF1* | 0.4059 | 0.2801 |
| FH | AUROC | 0.5750 | 0.5077 |
| | AUPRC | 0.4190 | 0.3342 |
| | MaAvg F1 | 0.5703 | 0.4923 |
| | *NegF1* | 0.7182 | 0.7745 |
| | *PosF1* | 0.4223 | 0.2101 |

Table 7: Performance of supervised-tuned models on raw discharge notes (non-summarized). Results for BoW+SVM and Hier-transformer across hospitals are reported.